\documentclass[sigconf]{acmart}

\usepackage{listings}
\usepackage[font=small]{caption}
\usepackage[subrefformat=parens]{subcaption}
\usepackage{array}
\usepackage{tabularx} 
\usepackage{balance}

\newlength\widest
\makeatletter
\NewEnviron{ldescription}{%
  \vbox{%
    \global\setlength\widest{0pt}%
    \def\item[##1]{%
      \settowidth\@tempdima{\textbf{##1}}%
      \ifdim\@tempdima>\widest\global\setlength\widest{\@tempdima}\fi%
    }%
    \setbox0=\hbox{\BODY}%
  }
  \begin{description}[
    leftmargin=\dimexpr\widest+0.5em\relax,
    labelindent=0pt,
    labelwidth=\widest]
  \BODY
  \end{description}%
}

\usepackage{url}
\PassOptionsToPackage{hyphens}{url}
\usepackage{hyperref}
\hypersetup{
  pdfborder={0 0 0},
}

\fancypagestyle{firstpagestyle}{
  \fancyhf{}
  \fancyfoot[C]{\if@ACM@printfolios\footnotesize\thepage\fi}
  \fancyhead[L]{\ACM@linecountL\@headfootfont xSIG 2023} 
  \fancyhead[R]{\ACM@linecountR}
}

\setcopyright{none}

\author{Limin Wang}
\email{lwang@eidos.ic.i.u-tokyo.ac.jp}
\affiliation{
  \institution{The University of Tokyo}
  \city{Tokyo}
  \country{Japan}
}

\author{Masatoshi Hanai}
\email{hanai@ds.itc.u-tokyo.ac.jp}
\affiliation{
  \institution{The University of Tokyo}
  \city{Tokyo}
  \country{Japan}
}

\author{Toyotaro Suzumura}
\email{suzumura@ds.itc.u-tokyo.ac.jp}
\affiliation{
  \institution{The University of Tokyo}
  \city{Tokyo}
  \country{Japan}
}

\author{Shun Takashige}
\email{shige@eidos.ic.i.u-tokyo.ac.jp}
\affiliation{
  \institution{The University of Tokyo}
   \city{Tokyo}
  \country{Japan}
}

\author{Kenjiro Taura}
\email{tau@eidos.ic.i.u-tokyo.ac.jp}
\affiliation{
  \institution{The University of Tokyo}
    \city{Tokyo}
  \country{Japan}
}

\renewcommand\footnotetextcopyrightpermission[1]{}

\begin{document}


\fancyfoot[C]{\if@ACM@printfolios\footnotesize \\ \thepage\fi}
\fancyhead[L]{\ACM@linecountL\@headfootfont xSIG 2023} 
\fancyhead[R]{\ACM@linecountR}
\fancypagestyle{firstpagestyle}{
  \fancyhf{}
  \fancyfoot[C]{\if@ACM@printfolios\footnotesize \\ \thepage\fi}
  \fancyhead[L]{\ACM@linecountL\@headfootfont xSIG 2023} 
  \fancyhead[R]{\ACM@linecountR}
}

\title{%
  On Data Imbalance in Molecular Property Prediction with Pre-training}

\begin{abstract}
    Revealing and analyzing the various properties of materials is an essential and critical issue in the development of materials, including batteries, semiconductors, catalysts, and pharmaceuticals.
    Traditionally, these properties have been determined through theoretical calculations and simulations.
    However, it is not practical to perform such calculations on every single candidate material.
    Recently, a combination method of the theoretical calculation and machine learning has emerged, that involves training machine learning models on a subset of theoretical calculation results to construct a surrogate model that can be applied to the remaining materials.
    On the other hand, a technique called pre-training is used to improve the accuracy of machine learning models.
    Pre-training involves training the model on pretext task, which is different from the target task, before training the model on the target task.
    This process aims to extract the input data features, stabilizing the learning process and improving its accuracy.
    However, in the case of molecular property prediction, there is a strong imbalance in the distribution of input data and features, which may lead to biased learning towards frequently occurring data during pre-training.
    In this study, we propose an effective pre-training method that addresses the imbalance in input data. We aim to improve the final accuracy by modifying the loss function of the existing representative pre-training method, node masking, to compensate the imbalance. We have investigated and assessed the impact of our proposed imbalance compensation on pre-training and the final prediction accuracy through experiments and evaluations using benchmark of molecular property prediction models.

\end{abstract}

\maketitle


\section{Introduction}
In recent years, there has been a growing trend in applying computational science and machine learning to material development and research, known as materials informatics.
In particular, understanding and analyzing the various properties of materials are a critical and essential issue in the development of materials such as batteries, semiconductors, catalysts, and pharmaceuticals. 
For analyzing material properties, it is not realistic to make and verify each candidate material one by one.
Therefore, in many cases, theoretical calculations are used to calculate and predict the target molecular properties.
The material property is typically calculated using molecular and crystal structures.

Recently, research on approaches using machine learning has been active.
It aims to construct a surrogate model that can be applied to other unknown materials by learning from some of the material properties obtained by theoretical calculations as the training data.
It difficult to calculate all candidates theoretically, because the number of candidate materials is countless, and the computational cost for some materials can be significantly high.
For example, computing property values often require an exponential order of computation for system size, and in some cases, large-scale computing resources such as supercomputers are necessary.

In machine-learning-based material property prediction, molecular data is represented as a graph structure and treated as input to Graph Neural Networks (GNN).
Moreover, in recent years, an approach using pre-training to machine-learning-based property prediction has gained attention to improve accuracy and stabilize learning.
Pre-training is a method of solving a pretext task first and then solving the downstream task by fine-tuning.
The effectiveness of pre-training has been widely demonstrated in fields such as natural language processing~\cite{BERT, brown2020language} and computer vision~\cite{he2020momentum, chen2020simple}.
By pre-training models, it is possible to learn information that may not be captured by downstream tasks alone, allowing for more comprehensive learning. 
In recent years, various studies have been conducted on the use of pre-training in material property prediction~\cite{Strategies, Does, GraphMVP, Unified2D-3D}.

While various studies have been conducted on GNN-based property prediction models using pre-training, this study aims to investigate and solve the problem of data imbalance.
Specifically, we aim to construct a model that performs effective pre-training while addressing the data imbalance in molecular property prediction.
Here we focus on the imbalance of element features among molecular data.
Element features belong to node features in molecular graphs and are used in pre-training on nodes. 
However, the frequency distribution of elements is not equal for all elements, and in organic compound molecules, hydrogen and carbon are overwhelmingly more frequent than others.
When there is an imbalance in the frequency of element features, it is expected that the imbalance can cause a problem of a biased learning towards abundant data, affecting the pre-training.
Furthermore, in property prediction, less frequent elements tend to have unique properties compared to the frequent elements, so if pre-training on nodes is biased, the model may miss the essential information, and the accuracy of predicting the target property after fine tuning may also deteriorate.
Thus in this research, we aim to propose a better pre-training method that solves the problem caused by the imbalance of element features.

In this study, we use a pretext task of reconstructing masked features of nodes (especially elements) during pre-training.
This task is surprisingly powerful in natural language processing.
This is also used in a lot of models of molecular property prediction and its effectiveness has been demonstrated.

Our contributions are as follows:
\begin{enumerate}
    \item We investigate and assess the data imbalance of features on Open Graph Benchmark~\cite{OGB-LSC}, which is the representative benchmark of GNN molecular property prediction.
    \item We propose the method of compensating imbalance by modifying loss function during pre-training on the existing pretext task: reconstructing masked features of nodes.
    \item We implement our proposal method on the existing SOTA model, Graphormer, and experimentally evaluate the impact of our proposed imbalance compensation on pre-training and the final prediction accuracy.
\end{enumerate}

\section{Background and Related Work \label{sec:background}}
\subsection{Molecular Property Prediction}
Molecular property prediction is to predict properties of a molecule from a molecular graph, which represents molecular structure, using the fact that the molecular and crystal structure of a substance can be represented as a graph.
The properties to be predicted vary from molecular shape, reactivity, reaction with electromagnetic fields, etc.
The reason for predicting properties from molecular structures is that they are relatively easy to obtain compared to complex properties, and there is a large amount of data available in open databases.
This prediction enables a wide range of applications, including the prediction of properties for the development of organic photovaltaic devices and the discovery of new drugs.

Molecular properties can be calculated for each substance by theoretical calculations and simulations such as density functional theory (DFT) ~\cite{burke2012perspective}, and are applied to material development by comparing and analysing properties of various substances.
However, it is not realistic to carry out theoretical calculations for all candidate materials in materials development using physical properties.
The number of candidate materials is often enormous, and in many cases large computational resources are required for theoretical calculations for each material.
Supercomputers are typically used for some properties that are computationally demanding on the exponential order of magnitude with respect to the size of the system.

In order to address these problems, machine-learning-based prediction of molecular properties has attracted much attention in recent years.
The main objective is to construct surrogate models, in which theoretical calculations are replaced by machine learning.
In other words, only part of the total data is subjected to theoretical calculations, and machine learning is used to construct a prediction model for the remaining data, using the calculated results as the training data.

\medskip
\noindent\underline{\textbf{Task: Open Graph Benchmark}}\\  
The Open Graph Benchmark (OGB) is an open-source package for comparing GNNs with indices, including large datasets of various tasks, by Hu et al~\cite{OGB-LSC}.
In this study, we use one of the various OGB tasks, PCQM4Mv2, a molecular property prediction task and its dataset.
In PCQM4Mv2, the task is to predict the HOMO-LUMO energy gap calculated by DFT.
For the HOMO-LUMO energy gap, HOMO refers to the highest energy of the molecular orbitals occupied by electrons (Highest Occupied Molecular Orbital) and LUMO refers to the lowest energy of the molecular orbitals not occupied by electrons (Lowest Unoccupied Molecular Orbital).
The HOMO-LUMO energy gap means the difference in energy between these two orbitals, and is one of the most fundamental and practical quantum chemical properties of molecules since it is related to their reactivity, photoexcitation and charge transport.
The task is a regression task to predict molecular properties in the form of scalar values from molecular structures in the form of graphs.
The Mean Absolute Error (MAE) of the HOMO-LUMO energy gap predicted by the model is used as the evaluation metric.

\subsection{Pre-training \label{sec:pre-train}}
This subsection describes pre-training.
Pre-training is often used in conjunction with fine-tuning.
Pre-training refers to first using a task different from the the target task (called the downstream task) in order to improve effectiveness in the target task.
By learning with a task different from the downstream task, more information can be learned from the graph and a better representation, which reflects more information from the graph, can be obtained to improve the performance in the downstream task.
The task used during pre-training is called a pretext task.
Fine tuning is done after pre-training.
It uses a model that has already been trained on the pretext task, and then is trained again on the downstream task, so that the learned model can be adjusted to perform well the downstream task.
Pre-training is used in many high-performance models in a variety of fields~\cite{BERT,brown2020language,wei2021finetuned,he2020momentum,chen2020simple}.

Among pre-training, self-supervised learning (SSL) pre-training has achieved good results in fields such as natural language processing~\cite{BERT, brown2020language} and computer vision~\cite{he2020momentum, chen2020simple}.
In self-supervised learning, labels to be used for learning are generated from the data itself.
One method of pre-training that uses self-supervised learning is pre-training by masking.

\medskip
\noindent\underline{\textbf{Pre-training by Masking}}\\  
In pre-training by masking, first some of the data in the input is masked.
The pretext task is to restore the masked data.
Masking is achieved by inserting a value indicating "being masked" instead of the original value.
"Pre-training by masking" is used in BERT~\cite{BERT} in natural language processing and has been very effective, so it has been applied to graphs as well.
When "pre-training by masking' is applied to graphs, the node and edge features of the input graph information can be randomly masked~\cite{Strategies, Does, Unified2D-3D}.
This technique has been shown to be particularly effective with the graph which contains abundant information in scientific fields, such as molecular graphs~\cite{Strategies}.
For example, in molecular graphs, the feature of a node represents the properties of each atom, and the task of masking and restoring it is expected to allow models to learn simple chemical regularities such as valence and more complex chemical phenomenon such as electronic and steric properties.

\subsection{Compensation of Data Imbalance \label{sec:vs_imbalance}}
Common open datasets in natural language processing or computer vision are often imbalanced, i.e. labels and features are not evenly distributed.
If a dataset has data imbalance, training can be biased towards to labels which exist a lot in the dataset, and for labels which exist only a few, the accuracy of the trained model can be significantly low ~\cite{cao2019learning, cui2019class, huang2019deep}.
Therefore, many methods have been proposed to address data imbalance.

In general, methods for dealing with data imbalance can be divided into two: one is "processing on data" and the other is "processing on models"~\cite{DeepImbalancedRegression, cao2019learning}.
The first, "processing on data" method, is a method of over-sampling minority labels or under-sampling majority labels.
This is also known as re-sampling.
Examples of this technique include ~\cite{chawla2002smote, garcia2009evolutionary, he2008adasyn}.
The second method of 'processing on model' refers to adjusting the loss function or weighting the loss function to compensate for data imbalance.
This is also known as re-weighting.
Examples of this method include ~\cite{cao2019learning, cui2019class, huang2019deep, phan2020resolving, aurelio2019learning}.

This study focuses on data imbalance in molecular property prediction and attempts to adapt the above general methods of compensating imbalance.

\section{Problem \label{sec:problem}}
The goal of this study is to predict the HOMO-LUMO energy gap.
And we work on improving the accuracy of the prediction by using pre-training.
Here, we utilize pre-training on node features.
Node feature involves the problem of data imbalance or biased distribution frequency, especially the elements of the atoms in the molecular graph.
In this study, we address this imbalance problem and work on better pre-training for molecular property prediction.

\subsection{Situation of Data Imbalnace}
Data imbalance is a common problem.
It is known that many distributions of data in the natural world follow the power law~\cite{reed2001pareto, newman2005power}.
The power law states that a function $Y$ of the number of data can be expressed, using some variable $x$, by the following equation.
\begin{align*}
Y = C x^{\alpha}
\end{align*}
This means that the data is distributed in proportion to a power of $x$.
Examples that obey the power law include the population of a city, the frequency of words, people's income, etc., as well as data on natural phenomena such as the scale of earthquakes or the size of a grain of sand.

This kind of data imbalance also exists in the field of molecular properties.
Some of examples of the data imbalance problem in molecular properties are the bias of the molecular properties themselves and the bias in the frequency of features of atoms (= nodes of molecular graphs).
In particular, the frequency of elemental types is unequal for atoms in organic compound molecules.
For example, Table~\ref{tab:atom_num} shows the elements and their frequency in a molecule which is expressed  as "COc1cc(OC)cccc1/C=C/N(C(=O)C)C" in SMILES (Simplified Molecular Input Line Entry System) notation.
Here, please note that hydrogen is omitted.
From Table~\ref{tab:atom_num}, carbon is the most abundant element, compared to nitrogen and oxygen which are less frequent, and this means that the elements are not equally distributed.
This bias in the frequency of elements in a molecule can be also observed in other organic compound molecules.

\begin{table}[]
    \centering
    \begin{tabular}{c|c}
        \hline
        Atomic Number / Element & Number (Pieces) \\
        \hline
        6 / C & 13 \\
        7 / N & 1 \\
        8 / O & 3 \\
        \hline
    \end{tabular}
    \caption{Elements and their existing numbers in molecule \\ "COc1cc(OC)ccc1/C=C/N(C(=O)C)C"}
    \label{tab:atom_num}
\end{table}

\begin{figure}
    \centering
    \includegraphics[width=\linewidth]{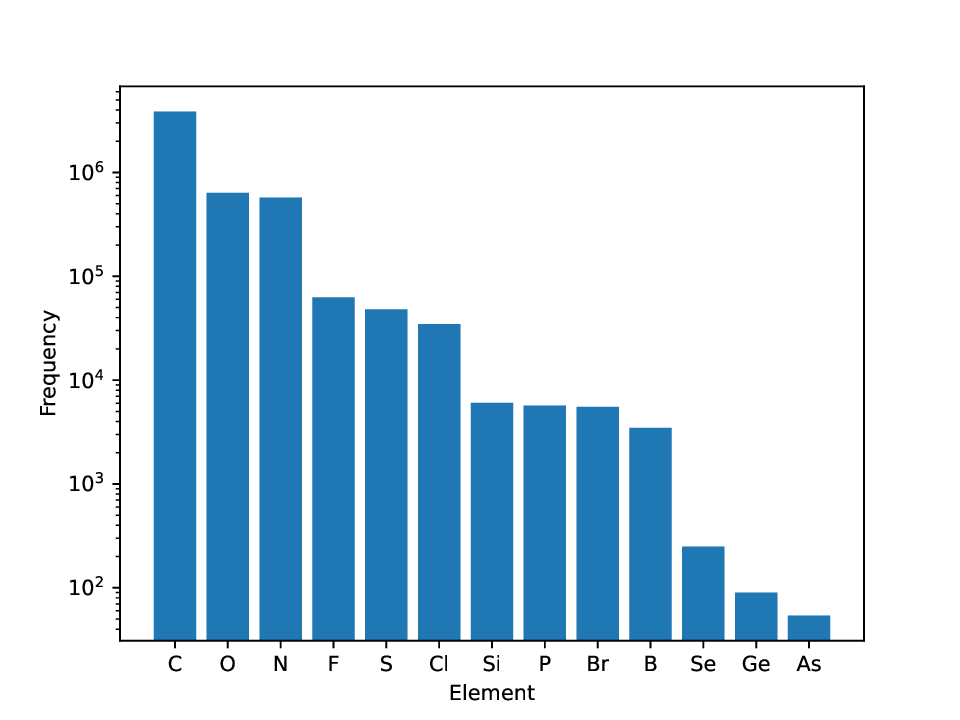}
    \caption{The distribution of elements in OGB-LSC's PCQM4Mv2}
    \label{fig:distribution_node_log}
\end{figure}

Furthermore, not only at the level of individual molecules, but also in the entire dataset, there exists a bias towards certain elements.
Specifically, in the PCQM4Mv2 dataset, which is a dataset of organic compound molecules within the OGB-LSC dataset, the distribution of element types of all nodes of all the graphs is shown in Figure~\ref{fig:distribution_node_log}.
The vertical axis is on a logarithmic scale.

Regarding the overall distribution, we can observe that C (carbon) is the most abundant element by a large margin, with its frequency being nearly 10 times that of the second most abundant element (O, oxygen).
What's more, even among the less frequent elements, there are differences of an order of magnitude in their frequency.
For instance, the frequency of O is approximately 10 times that of S (sulfur), and the frequency of S is 10 times that of P (phosphorus).
That is, the frequency of C is approximately 1000 times that of P.

Thus, there exists a significant imbalance in the data frequency of node elements, and it seems that this imbalance also follows a power law.
Furthermore, not only in the PCQM4Mv2 dataset, but also in many other datasets, it has been demonstrated that the number of carbons is much greater than that of other elements, and there are differences even within other elements.



\subsection{Impact of Data Imbalance on Pre-training in Molecular Property Prediction}
The data imbalance discussed above can cause a problem in pre-training of molecular property prediction.
For the rationale for this, Sun et al.~\cite{Does} argues that in the pretext task, if there is no diversity in the data used for training, the model quickly acquires generalisation capabilities, so that the model can learn less information than in the case of where there exists diversity in the data.
In the situation of this study, the data imbalance of elements of nodes results in a large imbalance in the feature which is used in the pretext task.
In other words, it is similar to the situation where almost only the "frequent elements" are present, which is close to the situation of no data diversity.
Therefore, we can consider that the data imbalance can reduce the information which the model can learn through pre-training, and also the effect of pre-training.
Furthermore, in property prediction, less frequent elements tend to have unique properties compared to the frequent elements, so if pre-training on nodes is biased, the model may miss the essential information, and the accuracy of predicting the target property after fine tuning may also deteriorate.

Additionally, we would also like to discuss the difference between this case of molecular property prediction and the model "BERT"~\cite{BERT} for natural language processing, both of which leverage pre-training by masking in common.
In natural language processing, there exists a bias in the distribution of word frequencies (~\cite{reed2001pareto, newman2005power}).
However, we could not find any existing studies that address this bias.
One possible reason why natural language processing does not need to address this word distribution bias is that languages have Distributional Hypothesis (~\cite{harris1954distributional, sahlgren2008distributional, sinha2021masked}).
Distributional Hypothesis is the idea that words with similar distributional frequencies in a context also have similar meanings, i.e. the distributional frequencies of words are correlated with the meaning of the words themselves.
Therefore, it is considered that the bias in the frequency of occurrence of words itself is meaningful so it will not cause a problem in pre-training.
In contrast, in the case of molecular property prediction, molecular properties are determined by the composition and type of elements, and the properties of elements should not depend on the frequency of occurrence of the elements.
Therefore, it is considered that data imbalance can cause a problem in the field of molecular property prediction.

\section{Proposed Method \label{sec:method}}
\subsection{Overview}
In this study, we propose a method for better pre-training with addressing the data imbalance problem discussed above in Section~\ref{sec:problem}.
We first pre-train a model with a task of "reconstruction of masked nodes in the graph as input" as the pretext task, and then we fine tune the model on the target molecular properties.
During the pre-training, we have the model compensate the data imbalance while solving the pretext task.
In this case, we use re-weighting by weighting the loss function to compensate the data imbalance.
In other words, small weights are applied to the data that exist in large numbers and large weights are applied to the data that exist in small numbers.
This ensures that data with large weights are more strongly reflected in the learning, leading to a compensation of the data imbalance.



\subsection{Pre-training and Fine Tuning}
This section describes the pre-training method for the material properties prediction which is used in the proposal.

In the proposal, we adopt the pre-training using node features, in particular we adopt the pretext task of reconstructing masked node features in the "pre-training by masking" described above in Section~\ref{sec:pre-train}.
In this case, we only mask the atomic number out of the nine features in each node, which we have the model reconstruct.
The nodes to mask is randomly selected in the graph at the proportion we set as a hyper parameter.
Since the node features, including the atomic numbers, are discrete values, the task of reconstructing the masked features is a classification task.
After the pre-training with this pretext task, we fine tune the model with the downstream task of predicting the HOMO-LUMO energy gap.






\subsection{Model}
In this study, the architecture of the model we use is based on Graphormer~\cite{Graphormer}.
The Encoder adopts Graphormer, and the output layer is a linear layer, which outputs the target molecular properties (in this case, HOMO-LUMO energy gap) in the downstream task and node features in the pretext task.
As for loss functions, we use the Mean Absolute Error (MAE) when predicting the HOMO-LUMO energy gap and the cross entropy loss when predicting the node.



\subsection{Proposal: Compensation of Imbalance by Loss Function during Pre-training}
In this study, we propose to compensate the data imbalance during pre-training.
As mentioned above in Section~\ref{sec:vs_imbalance}, there are two different methods for compensating data imbalance: re-sampling and re-weighting.
Re-sampling is a technique that increases the number of rare data or reduces the number of frequent data in the dataset.
In the case of molecules, however, re-sampling is difficult when dealing with the frequency imbalance of elements.
This is because, as can be seen from Figure~\ref{fig:distribution_node_log}, the differences in frequency between elements are very large.
For example, there are pairs of elements whose frequencies differ by more than 1000 times.
It is practically difficult to increase the number of rare elements data by 1000 times.
In addition, each molecule is inherently biased (especially carbon, which is very frequent).
Suppose that you want to increase the number of rare elements, you need to increase the number of molecular graphs which contains that element.
But those molecular graphs may contain frequent elements (such as carbon) that already have more data than that rare element, so you may not be able to only increase the number of rare elements, but also end up to increase the number of frequent elements much more.
That is, it is difficult to compensate data imbalance using re-sampling.

Therefore, in this proposal we use re-weighting to compensate the data imbalance.
This is a method of weighting the loss function.
We apply a smaller weight to the loss on the data that exist in large amounts and a larger weight to the loss on the data that exist in small amounts, so that the relatively small amount of data is reflected more strongly in the learning process.
By doing this, we aim to have the model learn not only from frequent data, but also from rare data.
In this proposal we apply re-weighting as a method for compensating data imbalance during pre-training.
That is, in the pre-training method by node masking described in the previous section, the model is trained with the weighted loss function.

In the original node masking pretext task, the cross entropy loss is used as the loss function.
The cross entropy loss is a loss function often used in classification tasks~\cite{wu2020recent}.
In the case of a $C$ class classification task, given some data $i$, suppose that the one-hot vector representing which class the data $i$ belongs to is $t_i$ (its size is $C$), and the result predicted by the classifier model (how likely it falls into each class) is $p_i$ (its size is $C$).
Then the cross entropy loss of the classification of that data $i$ is $l_i$, where $l_i$ is represented by Equation~\ref{eq:ce_i}~\cite{wang2020comprehensive, janocha2017loss, phan2020resolving, aurelio2019learning}.




\begin{equation}
\label{eq:ce_i}
    l_i = - ~ t_i \log{p_i}
\end{equation}

The cross entropy loss function, by which we evaluate the performance of the classifier, takes either of sum, or average, of the cross entropy loss of the total data, and can be represented by Equation~\ref{eq:ce} when the total number of data is $N$.


\begin{equation}
\label{eq:ce}
    L = 
    \left\{
    \begin{alignedat}{2}
        & \sum_i^N l_i    &\quad&    \text{(sum)}  \\
        & \frac{1}{N} \sum_i^N l_i  &     &    \text{(mean)}
    \end{alignedat}
    \right.
\end{equation}

For this cross entropy loss function, weights are assigned according to the number of data per class.
Let us suppose that the labels corresponding to the $C$ classes are ${y_1, y_2, y_3, ... , y_C}$, the weights to be applied to each class are ${w_1, w_2, ... , w_C}$, and let $W$ (with size $C$) be the vector of weights, to which all the weights ${w_1, w_2, ..., w_C}$ are put together.
Then we can describe the weighted cross entropy loss function as Equation~\ref{eq:wce_i},~\ref{eq:wce}.


\begin{equation}
\label{eq:wce_i}
    l_i = - ~ W ~ t_i \log{p_i}
\end{equation}

\begin{equation}
\label{eq:wce}
    L = 
    \left\{
    \begin{alignedat}{2}
        & \sum_i^N l_i    &\quad&    \text{(sum)}  \\
        & \frac{1}{\sum_i^N W t_i} \sum_i^N l_i  &     &    \text{(mean)}
    \end{alignedat}
    \right.
\end{equation}
This allows to reflect the loss more on classes with large weights.
(Note that the case with no weights (Equation~\ref{eq:ce}) is equivalent to the case where the weights ${w_1, w_2, w_3, ... , w_C}$ are all equal to $1$).
To counteract the imbalance in the number of data, larger weights should be applied to data of less frequency.
For instance, \cite{aurelio2019learning, phan2020resolving} uses the reciprocal of the number of data and so on as the weights to compensate the data imbalance.


In our proposal method, we adopt the following three ways to calculate the weight which is applied to each class.
\begin{enumerate}
    \item The weight is the reciprocal of the proportion of the number of data.
    \item The weight is "$1 ~ -$ proportion of the number of data".
    \item The weight is $-$ log times the proportion of the number of data.
\end{enumerate}
When the number of data in class $C_m$ out of $C$ classes is $N_m$, the total number of all data in all classes is $N_{all}$, and $r$ is the proportion of class $C_m$ in all data (Equation\ref{eq:class_ratio}), these three ways of weighting can be expressed in Equation~\ref{eq:3weight}.


\begin{equation}
\label{eq:3weight}
    w_m =
    \left\{
    \begin{alignedat}{2}
        & \frac{1}{r_m}  &\quad&    \text{(Reciprocal of propotion)}  \\
        & 1 - r_m &     &    \text{($1~-$propotion)}  \\
        & - \log{r_m} &     &    \text{($-$ log of propotion)}
    \end{alignedat}
    \right.
\end{equation}

\begin{equation}
\label{eq:class_ratio}
    r_m = \frac{N_m}{N_{all}}
\end{equation}

In the following, these three ways of calculating weights will be referred to as "Reciprocal", "Proportion" and "Log", in that order.
In addition, the case where no weights are applied, i.e. "the case where the weights ${w_1, w_2, w_3, ... , w_C}$ are all set to $1$" will be referred to as "No Weight" in the following.

Let us here consider the relationship between the respective weights and the strength of the data imbalance compensation.
The following is the derivative of the functions of the three weighting methods described in Section~\ref{sec:method}.
In the following, we use $w$ and $r$ as continuous variables for convenience, where $w$ and $r$ represents the weights ($w_i$) on the element $i$ and the proportions ($r_i$) of the element $i$ out of all atoms, respectively.



\begin{enumerate}
    \item "Reciprocal"
        \begin{equation}
            \frac{d}{dr}(w) = \frac{d}{dr}(\frac{1}{r}) = - \frac{1}{r^2}
        \end{equation}
    \item "Log"
        \begin{equation}
            \frac{d}{dr}(w) = \frac{d}{dr}(-\log{r}) = - \frac{1}{r}
        \end{equation}
    \item "Proportion"
        \begin{equation}
            \frac{d}{dr}(w) = \frac{d}{dr}(1 - r) = -1
        \end{equation}
    \item "No Weight"
        \begin{equation}
            \frac{d}{dr}(w) = \frac{d}{dr}(1) = 0
        \end{equation}
\end{enumerate}

Since the variable $r$ represents the proportion of the element among all atoms, $0 \le r \le 1$.
In this range of $r$, we can say that $- \frac{1}{r^2} \ll - \frac{1}{r} \ll - 1 < 0$, so that in the order of "reciprocal" $>$ "log" $>$ "percentage" $>$ "no weight", greater weight is applied to rare data and less weight to frequent data.
In other words, in this order, the compensation is stronger in the direction of counteracting "bias in the number of data".

In our proposal method, we use the average for cross entropy loss.
This is because, when we use the average, the loss is not affected by the scale of the weights.
For example, even if the weights of all classes are equally doubled, the value of loss will still be the same as when the weights are not doubled.
Therefore, the advantage is that we no longer need to consider the scale in the design of the weights.

Strictly speaking, when calculating the reciprocal or log of the proportion of the number of data, we add a very small number ($10^{-7}$) to the total number of data and then calculate the weights, in order to prevent the weights diverging for elements with $0$ of data.




\section{Evaluation \label{sec:evaluation}}
To evaluate the above proposed method, we used OGB-LSC's \\ PCQM4Mv2~\cite{OGB-LSC} as the benchmark dataset.

As described in Section~\ref{sec:method}, we first train a model with the task of masking nodes and restoring them as a pretext task in the pre-training.
The node to mask is chosen randomly for each graph at some proportion which we give as a hyper parameter.
Afterwards, we fine tune the model with the task of predicting the HOMO-LUMO energy gap for each molecule as a downstream task.
Due to the time constraints, we use one tenth of the total data from the benchmark PCQM4Mv2 for the evaluation.
On 4 GPUs, 100 epochs of pre-training and 100 epochs of fine tuning followed, each taking approximately 2.5 hours.

In this section, we evaluate the weighting methods described above for pre-training with weights on the loss function to deal with the data imbalance.
We evaluate them from the following three viewpoints.
\begin{enumerate}
    \item Difficulties of the pretext task
    \item Accuracy of the model after pre-training on the pretext task
    \item Accuracy of the model after fine tuning
\end{enumerate}

\subsection{Difficulties of the Pretext Task}
In this subsection, we investigate the difference between different weighting methods for the pre-training themselves.

\begin{figure}
    \centering
    \includegraphics[width=\linewidth]{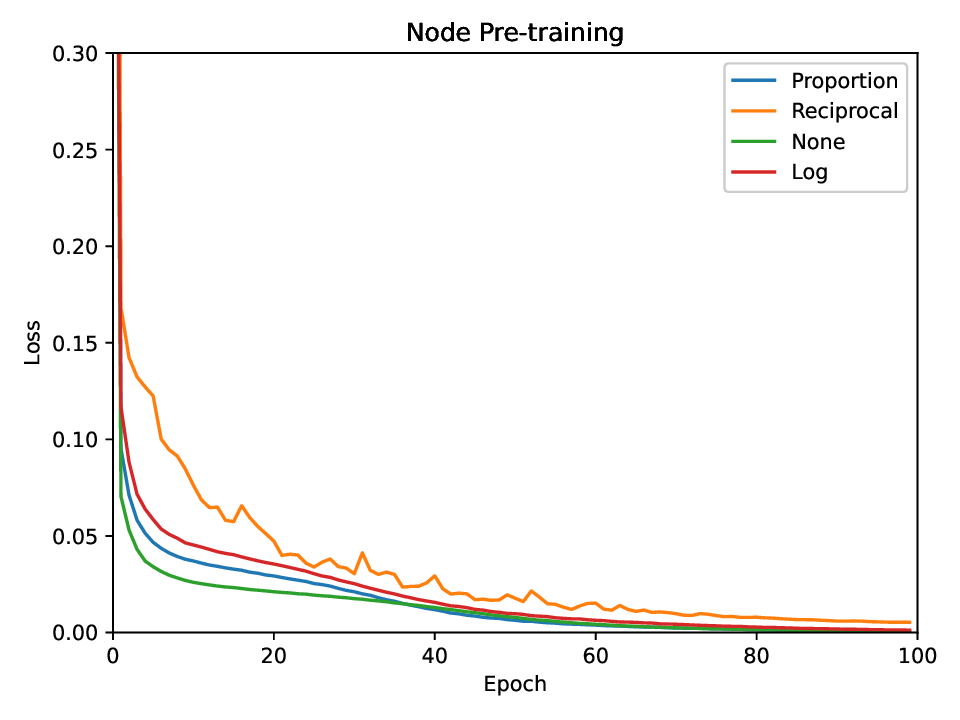}
    \caption{Loss transition during pre-training with 4 patterns of re-weighting}
    \label{fig:node-pretrain}
\end{figure}

We investigate how the loss undergoes its transition during the pre-training for four cases, including the three weighting methods proposed in the previous section, plus the unweighted case.
As mentioned above, cross entropy loss is used for the loss function in the pre-training.
Figure~\ref{fig:node-pretrain} shows the result of the transition of the loss during 100 epochs of pre-training, when we mask one node per a graph for the pretext task of reconstructing it.

As can be seen from Figure~\ref{fig:node-pretrain}, the fastest decrease/convergence of loss is for "None (No weight)", followed by "Proportion", then "Log", with "Reciprocal" having the slowest decrease in loss.
For example, in the case of "None (No weight)", the fastest convergence of loss, loss fell below 0.05 for the first time in the 4th epoch, whereas in the case of "reciprocal", the slowest convergence, loss fell below 0.05 for the first time in the 20th epoch.
When the convergence of the pre-training loss is fast, the pretext task can be regarded as easy, because the model is able to solve that pretext task earlier and more accurately than the task with which the loss converges slower.
Therefore, we can say that the rank of the difficulty of the pretext task is in the order of "Reciprocal" $>$ "Log" $>$ "Proportion" $>$ "No weight".
This order is consistent with the order of the strength of the imbalance compensation, i.e. the stronger the compensation is, the more difficult the pretext task is.

Figure~\ref{fig:node-pretrain} is the case when only one node is chosen to be mask per a graph, but even when we test the pre-training at other proportions to choose nodes to mask, the order of the difficulties of pretext task among the 4 types of re-weighting method remain the same.

According to Sun et al.~\cite{Does}, if the pretext task is too easy, it will not have a good effect on the downstream task.

\subsection{Accuracy after Pre-training on the Pretext Task}
In this subsection, we explore the compensation of the data imbalance by weighting.

We investigate the accuracy of node prediction by the model just after training on a pretext task of reconstructing the masked node for four patterns of weighting; the three weighting methods described in the previous section and the unweighted case.
The purpose of this experiment is to figure out how much the effect of data imbalance on the model is compensated by the weighted pre-training.
We use the pre-trained model which are only trained with the pretext task of reconstructing nodes and haven't been fine tuned yet on the HOMO-LUMO energy gap.
We have that model to predict the masked node feature in the same way as the pre-training, from the graphs of different molecules from the ones that are used in the pre-training.
In this case, we mask some atoms per molecular graphs in the same way and at the same proportion as in the pre-training, and then have the trained model predict elements of the masked atoms.
We examine the accuracy of this prediction.

The accuracy of the predictions is evaluated in terms of recall, following Phan et al.~\cite{phan2020resolving}
The recall is the percentage of the number of atoms for which the model correctly predicts the original element out of the number of masked atoms.
We examined not only the recall for each element that are masked, but also the recall for all elements.
Table~\ref{tab:recall} shows the results of the trained models of 4 patterns of re-weighting, all of which is just pre-trained for 100 epochs with the pretext task of reconstructing masked nodes, when only one node is chosen to be masked per a graph.
Since there are so many types of elements and, as mentioned above, the frequency of elements varies so greatly, here we summarise the recalls in four groups, grouped by elements with a similar frequency according to the number of data, in order to express the results more clearly.

\begin{table*}[t]
    \centering
    \begin{tabular}{c||c||c|c|c|c}
        \hline
        Element & Frequency & \begin{tabular}{c}
            Recall (\%) at \\ "no weight"
        \end{tabular} & \begin{tabular}{c}
            Recall (\%) at \\ "proportion"
        \end{tabular} & \begin{tabular}{c}
            Recall (\%) at \\ "log"
        \end{tabular} &  \begin{tabular}{c}
            Recall (\%) at \\ "reciprocal"
        \end{tabular} \\
        \hline
        C & 3856479 & 97.73 & 97.24 & 92.72 & 85.88 \\
        O,N & 1209103 & 51.46 & 61.39 & 66.91 & 53.18 \\
        F,S,Cl & 145488 & 16.53 & 39.12 & 49.33 & 60.46 \\
        Si,P,Br,B,Se & 20983 & 0.00 & 24.25 & 32.90 & 41.95 \\
        \hline
        Overall & 5232053 & 87.60 & 87.66 & 85.92 & 78.09 \\
        \hline
    \end{tabular}
    \caption{Recall of node prediction with 4 patterns of re-weighting on loss function}
    \label{tab:recall}
\end{table*}

Regarding the results, firstly we take a look at the relationship between the recall for each element group and the types of re-weighting to compensate the data imbalance.
About the 4 types of re-weighting, please note that the compensation of the data imbalance becomes stronger in the order of "no weight" $<$ "percentage" $<$ "log" $<$ "reciprocal".
As can see from the Table~\ref{tab:recall}, the recall for the group of rarest elements (Si, P, Br, B, Se), is the lowest on "no weight", followed by "percentage", "log", and the recall is the highest on "reciprocal".
That is, the recall for rare elements increases as the imbalance compensation becomes stronger.
On the other hand, we can see that the recall for frequent elements tends to decrease as the imbalance compensation becomes stronger.
From this, we can see that the compensation of the data imbalance prevents a bias in learning towards frequent elements.

However, the order of the overall recall is "no weight" $\approx$ "proportion" $>$ "log" $>$ "reciprocal".
This means that the recall decreases as the correction becomes stronger.
This is thought to be because the lower recall for overwhelmingly frequent elements is more significantly reflected in the overall recall due to the large number of frequent data.

\subsection{Accuracy after Fine Tuning}
In this subsection, we finally discuss the impact of re-weighting in pre-training on the accuracy of the model after fine tuning.
We compare the accuracy of the target property (HOMO-LUMO energy gap) prediction by the model after fine tuning with different re-weighting methods during pre-training.
We adopt Mean Absolute Error (MAE) of the HOMO-LUMO energy gap as the evaluation metric.
We conduct fine tuning for 100 epochs, and validation is performed once every 5 epochs within 100 epochs of fine tuning.
We use the minimum value of MAE for the validation data during the whole epochs to express the accuracy of the model.

To compare the MAE, we fine tune the pre-trained models of 4 types of weighting; the three ways of applying weights described in Section~\ref{sec:method} and also "no weight" for comparison.
All of the models are pre-trained for 100 epochs, and then fine tuned for 100 epochs.
For further reference, we also compare with a case where the model is without pre-training and trained only with the downstream task of the HOMO-LUMO energy gap for 100 epochs.
The following Table~\ref{tab:node_fine_tune} shows the results for all of these of when only one node is masked per graph.
Note that smaller values of MAE indicate better performance of the model.

\begin{table}[t]
    \centering
    \begin{tabular}{c|c}
        \hline
        Weights & min of validation MAE (eV) \\
        \hline
        No weight & 0.112 \\
        Proportion & 0.109 \\
        Log & 0.111 \\
        Reciprocal & 0.112 \\
        (Ref:no pre-training) & 0.112 \\
        \hline
    \end{tabular}
    \caption{Accuracy after fine tuning of the pre-trained model on 4 patterns of re-weighting, when one node is masked per a graph}
    \label{tab:node_fine_tune}
\end{table}

From Table~\ref{tab:node_fine_tune}, we can see that there is difference in the accuracy between the weighting methods, although the difference here is not very large.
The MAE value was the smallest for "Proportion" weight.


However, when we change the proportion of the masked nodes per a graph in the pretext task, we get different results.
In addition to the case of masking one node per a graph, whose result is shown in Table~\ref{tab:node_fine_tune}, we also examined the cases of masking 15\%, 30\%, 50\%, 80\% of all the nodes in a graph.
With all of these masking proportion, we compare the MAE of the fine tuned models on 4 weighting methods.
Table~\ref{tab:node_fine_tune_prop} shows the results for all of these cases.

\newcolumntype{Y}{>{\centering\arraybackslash}X}

\begin{table*}[t]
    \centering
    \begin{tabularx}{0.6\textwidth}{ c| *{4}{|Y} }
        \cline{2-5}
        \hline
        proportion of masked & \multicolumn{4}{c}{weights} \\
        \cline{2-5}
        nodes per a graph & No weight & Proportion & Log & Reciprocal \\
        \hline
        1 node & 0.112 & \textbf{0.109} & 0.111 & 0.112 \\
        15\% of nodes & \textbf{0.104} & 0.106 & 0.109 & 0.116 \\
        30\% of nodes & \textbf{0.102} & 0.108 & 0.108 & 0.111 \\
        50\% of nodes & \textbf{0.105} & 0.110 & 0.108 & 0.111 \\
        80\% of nodes & 0.107 & 0.108 & \textbf{0.106} & 0.108 \\
        \hline
    \end{tabularx}
    \caption{The proportion of masked nodes per a graph, and accuracy after fine tuning of the pre-trained model on 4 patterns of re-weighting}
    \label{tab:node_fine_tune_prop}
\end{table*}

As we can see from Table~\ref{tab:node_fine_tune_prop}, the results are different with different proportion of masked nodes per a graph.
For example, when the proportion of masked nodes is 30\%, the MAE is the smallest for "No weight", while the MAE is the smallest for "Log" weights when 80\% of nodes in a graph are masked.
From these result, we can say that re-weighting on the loss function during pre-training has influence on the accuracy of the fine tuned model.
Sometimes re-weighting improves the accuracy, but we couldn't observe that re-weighting always gives the positive effect.
We consider exploring this problem is one of the future work.

\section{Conclusion and Future Work \label{sec:conclusion}}
In this study, we designed a method for pre-training that considers data imbalance in molecular property prediction, and we experimentally evaluated the effect of compensating the data imbalance on pre-training and final prediction accuracy.
For pre-training, we used a pretext task of reconstructing the masked features of nodes (especially, their elements).
To address the imbalanced distribution of elements in the data, we applied small weights to frequent elements and large weights to infrequent elements, preventing the model from learning only from the frequent data and enabling it to learn from the infrequent data as well.
The strength of compensation of the data imbalance was determined by the weight applied to each element, and in this study we tried 3 compensation methods ("reciprocal", "log" and "proportion") and also the case of "no weight".
The compensation of the data imbalance becomes stronger in the order of "no weight" $<$ "percentage" $<$ "log" $<$ "reciprocal".
We found that as the compensation is stronger, the pretext task gets more difficult, and the biased learning toward the frequent data gets improved more.
On the final accuracy of the fine tuned model on the target property, we observed that the compensation of the data imbalance gives some influence on the model, and sometimes it improves the accuracy, but it doesn't seem that the strong compensation always works in the positive way.

As for the future prospects, we firstly aim to analyze the details using explainability techniques such as visualization of intermediate representations.
Then, we would like to propose new improvement methods based on the analysis.
Additionally, we also plan to try other methods that were not used this time.
Regarding the design of the loss function, we used a simple one this time, but we want to pursue more effective methods.
As for the pretext task of pre-training, we used the attribute masking technique at the node level called node mask this time, but there is room to try other tasks such as context prediction or graph-level tasks, and even tasks such as denoising, which predict the added noise correctly after adding noise as input.
Furthermore, we want to tackle other datasets besides the benchmark dataset used this time.
The OGB-LSC's PCQM4Mv2~\cite{OGB-LSC}, which we used in this study, was a dataset of organic compound molecules.
We want to apply it to other datasets of organic compounds and also inorganic benchmarks such as Open Catalyst~\cite{tran2022open} as well.




\section*{Acknowledgment}
This work is partially supported by "Advanced Research Infrastructure for Materials and Nanotechnology in Japan (ARIM)" of the Ministry of Education, Culture, Sports, Science and Technology (MEXT), and by JSPS KAKENHI Grant Number 22K17899.

\balance

\bibliographystyle{ACM-Reference-Format}
\bibliography{mainNotes}


\begin{thebibliography}{30}


\ifx \showCODEN    \undefined \def \showCODEN     #1{\unskip}     \fi
\ifx \showDOI      \undefined \def \showDOI       #1{#1}\fi
\ifx \showISBNx    \undefined \def \showISBNx     #1{\unskip}     \fi
\ifx \showISBNxiii \undefined \def \showISBNxiii  #1{\unskip}     \fi
\ifx \showISSN     \undefined \def \showISSN      #1{\unskip}     \fi
\ifx \showLCCN     \undefined \def \showLCCN      #1{\unskip}     \fi
\ifx \shownote     \undefined \def \shownote      #1{#1}          \fi
\ifx \showarticletitle \undefined \def \showarticletitle #1{#1}   \fi
\ifx \showURL      \undefined \def \showURL       {\relax}        \fi
\providecommand\bibfield[2]{#2}
\providecommand\bibinfo[2]{#2}
\providecommand\natexlab[1]{#1}
\providecommand\showeprint[2][]{arXiv:#2}

\bibitem[Aurelio et~al\mbox{.}(2019)]%
        {aurelio2019learning}
\bibfield{author}{\bibinfo{person}{Yuri~Sousa Aurelio},
  \bibinfo{person}{Gustavo~Matheus De~Almeida},
  \bibinfo{person}{Cristiano~Leite de Castro}, {and}
  \bibinfo{person}{Antonio~Padua Braga}.} \bibinfo{year}{2019}\natexlab{}.
\newblock \showarticletitle{Learning from imbalanced data sets with weighted
  cross-entropy function}.
\newblock \bibinfo{journal}{\emph{Neural processing letters}}
  \bibinfo{volume}{50} (\bibinfo{year}{2019}), \bibinfo{pages}{1937--1949}.
\newblock


\bibitem[Brown et~al\mbox{.}(2020)]%
        {brown2020language}
\bibfield{author}{\bibinfo{person}{Tom Brown}, \bibinfo{person}{Benjamin Mann},
  \bibinfo{person}{Nick Ryder}, \bibinfo{person}{Melanie Subbiah},
  \bibinfo{person}{Jared~D Kaplan}, \bibinfo{person}{Prafulla Dhariwal},
  \bibinfo{person}{Arvind Neelakantan}, \bibinfo{person}{Pranav Shyam},
  \bibinfo{person}{Girish Sastry}, \bibinfo{person}{Amanda Askell},
  {et~al\mbox{.}}} \bibinfo{year}{2020}\natexlab{}.
\newblock \showarticletitle{Language models are few-shot learners}.
\newblock \bibinfo{journal}{\emph{Advances in neural information processing
  systems}}  \bibinfo{volume}{33} (\bibinfo{year}{2020}),
  \bibinfo{pages}{1877--1901}.
\newblock


\bibitem[Burke(2012)]%
        {burke2012perspective}
\bibfield{author}{\bibinfo{person}{Kieron Burke}.}
  \bibinfo{year}{2012}\natexlab{}.
\newblock \showarticletitle{Perspective on density functional theory}.
\newblock \bibinfo{journal}{\emph{The Journal of chemical physics}}
  \bibinfo{volume}{136}, \bibinfo{number}{15} (\bibinfo{year}{2012}),
  \bibinfo{pages}{150901}.
\newblock


\bibitem[Cao et~al\mbox{.}(2019)]%
        {cao2019learning}
\bibfield{author}{\bibinfo{person}{Kaidi Cao}, \bibinfo{person}{Colin Wei},
  \bibinfo{person}{Adrien Gaidon}, \bibinfo{person}{Nikos Arechiga}, {and}
  \bibinfo{person}{Tengyu Ma}.} \bibinfo{year}{2019}\natexlab{}.
\newblock \showarticletitle{Learning Imbalanced Datasets with
  Label-Distribution-Aware Margin Loss}. \bibinfo{publisher}{arXiv}.
\newblock
\urldef\tempurl%
\url{https://doi.org/10.48550/ARXIV.1906.07413}
\showDOI{\tempurl}


\bibitem[Chawla et~al\mbox{.}(2002)]%
        {chawla2002smote}
\bibfield{author}{\bibinfo{person}{Nitesh~V Chawla}, \bibinfo{person}{Kevin~W
  Bowyer}, \bibinfo{person}{Lawrence~O Hall}, {and} \bibinfo{person}{W~Philip
  Kegelmeyer}.} \bibinfo{year}{2002}\natexlab{}.
\newblock \showarticletitle{SMOTE: synthetic minority over-sampling technique}.
\newblock \bibinfo{journal}{\emph{Journal of artificial intelligence research}}
   \bibinfo{volume}{16} (\bibinfo{year}{2002}), \bibinfo{pages}{321--357}.
\newblock


\bibitem[Chen et~al\mbox{.}(2020)]%
        {chen2020simple}
\bibfield{author}{\bibinfo{person}{Ting Chen}, \bibinfo{person}{Simon
  Kornblith}, \bibinfo{person}{Mohammad Norouzi}, {and}
  \bibinfo{person}{Geoffrey Hinton}.} \bibinfo{year}{2020}\natexlab{}.
\newblock \showarticletitle{A simple framework for contrastive learning of
  visual representations}. In \bibinfo{booktitle}{\emph{International
  conference on machine learning}}. PMLR, \bibinfo{pages}{1597--1607}.
\newblock


\bibitem[Cui et~al\mbox{.}(2019)]%
        {cui2019class}
\bibfield{author}{\bibinfo{person}{Yin Cui}, \bibinfo{person}{Menglin Jia},
  \bibinfo{person}{Tsung-Yi Lin}, \bibinfo{person}{Yang Song}, {and}
  \bibinfo{person}{Serge Belongie}.} \bibinfo{year}{2019}\natexlab{}.
\newblock \showarticletitle{Class-balanced loss based on effective number of
  samples}. In \bibinfo{booktitle}{\emph{Proceedings of the IEEE/CVF conference
  on computer vision and pattern recognition}}. \bibinfo{pages}{9268--9277}.
\newblock


\bibitem[Devlin et~al\mbox{.}(2018)]%
        {BERT}
\bibfield{author}{\bibinfo{person}{Jacob Devlin}, \bibinfo{person}{Ming-Wei
  Chang}, \bibinfo{person}{Kenton Lee}, {and} \bibinfo{person}{Kristina
  Toutanova}.} \bibinfo{year}{2018}\natexlab{}.
\newblock \showarticletitle{BERT: Pre-training of Deep Bidirectional
  Transformers for Language Understanding}. \bibinfo{publisher}{arXiv}.
\newblock
\urldef\tempurl%
\url{https://doi.org/10.48550/ARXIV.1810.04805}
\showDOI{\tempurl}


\bibitem[Garc{\'\i}a and Herrera(2009)]%
        {garcia2009evolutionary}
\bibfield{author}{\bibinfo{person}{Salvador Garc{\'\i}a} {and}
  \bibinfo{person}{Francisco Herrera}.} \bibinfo{year}{2009}\natexlab{}.
\newblock \showarticletitle{Evolutionary Undersampling for Classification with
  Imbalanced Datasets: Proposals and Taxonomy}.
\newblock \bibinfo{journal}{\emph{Evolutionary Computation}}
  \bibinfo{volume}{17}, \bibinfo{number}{3} (\bibinfo{year}{2009}),
  \bibinfo{pages}{275--306}.
\newblock


\bibitem[Harris(1954)]%
        {harris1954distributional}
\bibfield{author}{\bibinfo{person}{Zellig~S Harris}.}
  \bibinfo{year}{1954}\natexlab{}.
\newblock \showarticletitle{Distributional structure}.
\newblock \bibinfo{journal}{\emph{Word}} \bibinfo{volume}{10},
  \bibinfo{number}{2-3} (\bibinfo{year}{1954}), \bibinfo{pages}{146--162}.
\newblock


\bibitem[He et~al\mbox{.}(2008)]%
        {he2008adasyn}
\bibfield{author}{\bibinfo{person}{Haibo He}, \bibinfo{person}{Yang Bai},
  \bibinfo{person}{Edwardo~A Garcia}, {and} \bibinfo{person}{Shutao Li}.}
  \bibinfo{year}{2008}\natexlab{}.
\newblock \showarticletitle{ADASYN: Adaptive synthetic sampling approach for
  imbalanced learning}. In \bibinfo{booktitle}{\emph{2008 IEEE international
  joint conference on neural networks (IEEE world congress on computational
  intelligence)}}. IEEE, \bibinfo{pages}{1322--1328}.
\newblock


\bibitem[He et~al\mbox{.}(2020)]%
        {he2020momentum}
\bibfield{author}{\bibinfo{person}{Kaiming He}, \bibinfo{person}{Haoqi Fan},
  \bibinfo{person}{Yuxin Wu}, \bibinfo{person}{Saining Xie}, {and}
  \bibinfo{person}{Ross Girshick}.} \bibinfo{year}{2020}\natexlab{}.
\newblock \showarticletitle{Momentum contrast for unsupervised visual
  representation learning}. In \bibinfo{booktitle}{\emph{Proceedings of the
  IEEE/CVF conference on computer vision and pattern recognition}}.
  \bibinfo{pages}{9729--9738}.
\newblock


\bibitem[Hu et~al\mbox{.}(2021)]%
        {OGB-LSC}
\bibfield{author}{\bibinfo{person}{Weihua Hu}, \bibinfo{person}{Matthias Fey},
  \bibinfo{person}{Hongyu Ren}, \bibinfo{person}{Maho Nakata},
  \bibinfo{person}{Yuxiao Dong}, {and} \bibinfo{person}{Jure Leskovec}.}
  \bibinfo{year}{2021}\natexlab{}.
\newblock \showarticletitle{OGB-LSC: A Large-Scale Challenge for Machine
  Learning on Graphs}. \bibinfo{publisher}{arXiv}.
\newblock
\urldef\tempurl%
\url{https://doi.org/10.48550/ARXIV.2103.09430}
\showDOI{\tempurl}


\bibitem[Hu et~al\mbox{.}(2019)]%
        {Strategies}
\bibfield{author}{\bibinfo{person}{Weihua Hu}, \bibinfo{person}{Bowen Liu},
  \bibinfo{person}{Joseph Gomes}, \bibinfo{person}{Marinka Zitnik},
  \bibinfo{person}{Percy Liang}, \bibinfo{person}{Vijay Pande}, {and}
  \bibinfo{person}{Jure Leskovec}.} \bibinfo{year}{2019}\natexlab{}.
\newblock \showarticletitle{Strategies for Pre-training Graph Neural Networks}.
  \bibinfo{publisher}{arXiv}.
\newblock
\urldef\tempurl%
\url{https://doi.org/10.48550/ARXIV.1905.12265}
\showDOI{\tempurl}


\bibitem[Huang et~al\mbox{.}(2019)]%
        {huang2019deep}
\bibfield{author}{\bibinfo{person}{Chen Huang}, \bibinfo{person}{Yining Li},
  \bibinfo{person}{Chen~Change Loy}, {and} \bibinfo{person}{Xiaoou Tang}.}
  \bibinfo{year}{2019}\natexlab{}.
\newblock \showarticletitle{Deep imbalanced learning for face recognition and
  attribute prediction}.
\newblock \bibinfo{journal}{\emph{IEEE transactions on pattern analysis and
  machine intelligence}} \bibinfo{volume}{42}, \bibinfo{number}{11}
  (\bibinfo{year}{2019}), \bibinfo{pages}{2781--2794}.
\newblock


\bibitem[Janocha and Czarnecki(2017)]%
        {janocha2017loss}
\bibfield{author}{\bibinfo{person}{Katarzyna Janocha} {and}
  \bibinfo{person}{Wojciech~Marian Czarnecki}.}
  \bibinfo{year}{2017}\natexlab{}.
\newblock \showarticletitle{On loss functions for deep neural networks in
  classification}.
\newblock \bibinfo{journal}{\emph{arXiv preprint arXiv:1702.05659}}
  (\bibinfo{year}{2017}).
\newblock


\bibitem[Liu et~al\mbox{.}(2021)]%
        {GraphMVP}
\bibfield{author}{\bibinfo{person}{Shengchao Liu}, \bibinfo{person}{Hanchen
  Wang}, \bibinfo{person}{Weiyang Liu}, \bibinfo{person}{Joan Lasenby},
  \bibinfo{person}{Hongyu Guo}, {and} \bibinfo{person}{Jian Tang}.}
  \bibinfo{year}{2021}\natexlab{}.
\newblock \showarticletitle{Pre-training Molecular Graph Representation with 3D
  Geometry}. \bibinfo{publisher}{arXiv}.
\newblock
\urldef\tempurl%
\url{https://doi.org/10.48550/ARXIV.2110.07728}
\showDOI{\tempurl}


\bibitem[Newman(2005)]%
        {newman2005power}
\bibfield{author}{\bibinfo{person}{Mark~EJ Newman}.}
  \bibinfo{year}{2005}\natexlab{}.
\newblock \showarticletitle{Power laws, Pareto distributions and Zipf's law}.
\newblock \bibinfo{journal}{\emph{Contemporary physics}} \bibinfo{volume}{46},
  \bibinfo{number}{5} (\bibinfo{year}{2005}), \bibinfo{pages}{323--351}.
\newblock


\bibitem[Phan and Yamamoto(2020)]%
        {phan2020resolving}
\bibfield{author}{\bibinfo{person}{Trong~Huy Phan} {and}
  \bibinfo{person}{Kazuma Yamamoto}.} \bibinfo{year}{2020}\natexlab{}.
\newblock \showarticletitle{Resolving class imbalance in object detection with
  weighted cross entropy losses}.
\newblock \bibinfo{journal}{\emph{arXiv preprint arXiv:2006.01413}}
  (\bibinfo{year}{2020}).
\newblock


\bibitem[Reed(2001)]%
        {reed2001pareto}
\bibfield{author}{\bibinfo{person}{William~J Reed}.}
  \bibinfo{year}{2001}\natexlab{}.
\newblock \showarticletitle{The Pareto, Zipf and other power laws}.
\newblock \bibinfo{journal}{\emph{Economics letters}} \bibinfo{volume}{74},
  \bibinfo{number}{1} (\bibinfo{year}{2001}), \bibinfo{pages}{15--19}.
\newblock


\bibitem[Sahlgren(2008)]%
        {sahlgren2008distributional}
\bibfield{author}{\bibinfo{person}{Magnus Sahlgren}.}
  \bibinfo{year}{2008}\natexlab{}.
\newblock \showarticletitle{The distributional hypothesis}.
\newblock \bibinfo{journal}{\emph{Italian Journal of Disability Studies}}
  \bibinfo{volume}{20} (\bibinfo{year}{2008}), \bibinfo{pages}{33--53}.
\newblock


\bibitem[Sinha et~al\mbox{.}(2021)]%
        {sinha2021masked}
\bibfield{author}{\bibinfo{person}{Koustuv Sinha}, \bibinfo{person}{Robin Jia},
  \bibinfo{person}{Dieuwke Hupkes}, \bibinfo{person}{Joelle Pineau},
  \bibinfo{person}{Adina Williams}, {and} \bibinfo{person}{Douwe Kiela}.}
  \bibinfo{year}{2021}\natexlab{}.
\newblock \showarticletitle{Masked language modeling and the distributional
  hypothesis: Order word matters pre-training for little}.
\newblock \bibinfo{journal}{\emph{arXiv preprint arXiv:2104.06644}}
  (\bibinfo{year}{2021}).
\newblock


\bibitem[Sun et~al\mbox{.}(2022)]%
        {Does}
\bibfield{author}{\bibinfo{person}{Ruoxi Sun}, \bibinfo{person}{Hanjun Dai},
  {and} \bibinfo{person}{Adams~Wei Yu}.} \bibinfo{year}{2022}\natexlab{}.
\newblock \showarticletitle{Does GNN Pretraining Help Molecular
  Representation?} \bibinfo{publisher}{arXiv}.
\newblock
\urldef\tempurl%
\url{https://doi.org/10.48550/ARXIV.2207.06010}
\showDOI{\tempurl}


\bibitem[Tran et~al\mbox{.}(2022)]%
        {tran2022open}
\bibfield{author}{\bibinfo{person}{Richard Tran}, \bibinfo{person}{Janice Lan},
  \bibinfo{person}{Muhammed Shuaibi}, \bibinfo{person}{Siddharth Goyal},
  \bibinfo{person}{Brandon~M Wood}, \bibinfo{person}{Abhishek Das},
  \bibinfo{person}{Javier Heras-Domingo}, \bibinfo{person}{Adeesh Kolluru},
  \bibinfo{person}{Ammar Rizvi}, \bibinfo{person}{Nima Shoghi},
  {et~al\mbox{.}}} \bibinfo{year}{2022}\natexlab{}.
\newblock \showarticletitle{The open catalyst 2022 (OC22) dataset and
  challenges for oxide electrocatalysis}.
\newblock \bibinfo{journal}{\emph{arXiv preprint arXiv:2206.08917}}
  (\bibinfo{year}{2022}).
\newblock


\bibitem[Wang et~al\mbox{.}(2020)]%
        {wang2020comprehensive}
\bibfield{author}{\bibinfo{person}{Qi Wang}, \bibinfo{person}{Yue Ma},
  \bibinfo{person}{Kun Zhao}, {and} \bibinfo{person}{Yingjie Tian}.}
  \bibinfo{year}{2020}\natexlab{}.
\newblock \showarticletitle{A comprehensive survey of loss functions in machine
  learning}.
\newblock \bibinfo{journal}{\emph{Annals of Data Science}}
  (\bibinfo{year}{2020}), \bibinfo{pages}{1--26}.
\newblock


\bibitem[Wei et~al\mbox{.}(2021)]%
        {wei2021finetuned}
\bibfield{author}{\bibinfo{person}{Jason Wei}, \bibinfo{person}{Maarten Bosma},
  \bibinfo{person}{Vincent~Y Zhao}, \bibinfo{person}{Kelvin Guu},
  \bibinfo{person}{Adams~Wei Yu}, \bibinfo{person}{Brian Lester},
  \bibinfo{person}{Nan Du}, \bibinfo{person}{Andrew~M Dai}, {and}
  \bibinfo{person}{Quoc~V Le}.} \bibinfo{year}{2021}\natexlab{}.
\newblock \showarticletitle{Finetuned language models are zero-shot learners}.
\newblock \bibinfo{journal}{\emph{arXiv preprint arXiv:2109.01652}}
  (\bibinfo{year}{2021}).
\newblock


\bibitem[Wu et~al\mbox{.}(2020)]%
        {wu2020recent}
\bibfield{author}{\bibinfo{person}{Xiongwei Wu}, \bibinfo{person}{Doyen Sahoo},
  {and} \bibinfo{person}{Steven~CH Hoi}.} \bibinfo{year}{2020}\natexlab{}.
\newblock \showarticletitle{Recent advances in deep learning for object
  detection}.
\newblock \bibinfo{journal}{\emph{Neurocomputing}}  \bibinfo{volume}{396}
  (\bibinfo{year}{2020}), \bibinfo{pages}{39--64}.
\newblock


\bibitem[Yang et~al\mbox{.}(2021)]%
        {DeepImbalancedRegression}
\bibfield{author}{\bibinfo{person}{Yuzhe Yang}, \bibinfo{person}{Kaiwen Zha},
  \bibinfo{person}{Ying-Cong Chen}, \bibinfo{person}{Hao Wang}, {and}
  \bibinfo{person}{Dina Katabi}.} \bibinfo{year}{2021}\natexlab{}.
\newblock \showarticletitle{Delving into Deep Imbalanced Regression}.
  \bibinfo{publisher}{arXiv}.
\newblock
\urldef\tempurl%
\url{https://doi.org/10.48550/ARXIV.2102.09554}
\showDOI{\tempurl}


\bibitem[Ying et~al\mbox{.}(2021)]%
        {Graphormer}
\bibfield{author}{\bibinfo{person}{Chengxuan Ying}, \bibinfo{person}{Tianle
  Cai}, \bibinfo{person}{Shengjie Luo}, \bibinfo{person}{Shuxin Zheng},
  \bibinfo{person}{Guolin Ke}, \bibinfo{person}{Di He},
  \bibinfo{person}{Yanming Shen}, {and} \bibinfo{person}{Tie-Yan Liu}.}
  \bibinfo{year}{2021}\natexlab{}.
\newblock \showarticletitle{Do transformers really perform badly for graph
  representation?}
\newblock \bibinfo{journal}{\emph{Advances in Neural Information Processing
  Systems}}  \bibinfo{volume}{34}, \bibinfo{pages}{28877--28888}.
\newblock


\bibitem[Zhu et~al\mbox{.}(2022)]%
        {Unified2D-3D}
\bibfield{author}{\bibinfo{person}{Jinhua Zhu}, \bibinfo{person}{Yingce Xia},
  \bibinfo{person}{Lijun Wu}, \bibinfo{person}{Shufang Xie},
  \bibinfo{person}{Tao Qin}, \bibinfo{person}{Wengang Zhou},
  \bibinfo{person}{Houqiang Li}, {and} \bibinfo{person}{Tie-Yan Liu}.}
  \bibinfo{year}{2022}\natexlab{}.
\newblock \showarticletitle{Unified 2D and 3D Pre-Training of Molecular
  Representations}. In \bibinfo{booktitle}{\emph{Proceedings of the 28th ACM
  SIGKDD Conference on Knowledge Discovery and Data Mining}} (Washington DC,
  USA) \emph{(\bibinfo{series}{KDD '22})}. \bibinfo{publisher}{Association for
  Computing Machinery}, \bibinfo{address}{New York, NY, USA},
  \bibinfo{pages}{2626–2636}.
\newblock
\showISBNx{9781450393850}
\urldef\tempurl%
\url{https://doi.org/10.1145/3534678.3539368}
\showDOI{\tempurl}


\end{thebibliography}

\end{document}